\documentclass[journal]{IEEEtran}  
\usepackage{cite}
\usepackage{authblk}
\usepackage{amsmath}
\usepackage{amsfonts}
\usepackage{graphicx}
\usepackage{nohyperref}
\usepackage{url} 
\usepackage[ruled,longend]{algorithm2e}

\newcommand\blfootnote[1]{%
  \begingroup
  \renewcommand\thefootnote{}\footnote{#1}%
  \addtocounter{footnote}{-1}%
  \endgroup
}

\makeatletter
\def\bbordermatrix#1{\begingroup \m@th
  \@tempdima 4.75\p@
  \setbox\z@\vbox{%
    \def\cr{\crcr\noalign{\kern2\p@\global\let\cr\endline}}%
    \ialign{$##$\hfil\kern2\p@\kern\@tempdima&\thinspace\hfil$##$\hfil
      &&\quad\hfil$##$\hfil\crcr
      \omit\strut\hfil\crcr\noalign{\kern-\baselineskip}%
      #1\crcr\omit\strut\cr}}%
  \setbox\tw@\vbox{\unvcopy\z@\global\setbox\@ne\lastbox}%
  \setbox\tw@\hbox{\unhbox\@ne\unskip\global\setbox\@ne\lastbox}%
  \setbox\tw@\hbox{$\kern\wd\@ne\kern-\@tempdima\left[\kern-\wd\@ne
    \global\setbox\@ne\vbox{\box\@ne\kern2\p@}%
    \vcenter{\kern-\ht\@ne\unvbox\z@\kern-\baselineskip}\,\right]$}%
  \null\;\vbox{\kern\ht\@ne\box\tw@}\endgroup}
\makeatother
 
\title{\LARGE \bf
Universal Activation Function For Machine Learning
}

\author{Brosnan Yuen, Minh Tu Hoang, Xiaodai Dong, and Tao Lu}

\begin{document}

\maketitle

\begin{abstract}
\blfootnote{
This work was supported in part by the Nature Science and Engineering Research Council of Canada (NSERC) Discovery (Grant No. RGPIN-2020-05938 \& RGPIN-2018-03778),  and Threat Reduction Agency (DTRA) Thrust Area 7, Topic G18 (Grant No.GRANT12500317) and NVidia under GPU Grant program (\textit{Corresponding author: T. Lu.}). \\
B. Yuen, M. T. Hoang, X. Dong and T. Lu are with the
Department of Electrical and Computer Engineering, University of Victoria,
Victoria, BC, Canada (email: taolu@ece.uvic.ca).}

This article proposes a Universal Activation Function (UAF) that achieves near optimal performance in quantification, classification, and reinforcement learning (RL) problems.  For any given problem, the optimization algorithms are able to evolve the UAF to a suitable activation function by tuning the UAF's parameters. For the  CIFAR-10 classification and VGG-8, the UAF converges to the Mish like activation function, which has near optimal performance $F_{1} = 0.9017\pm0.0040$ when compared to other activation functions. For the quantification of simulated 9-gas mixtures in 30 dB signal-to-noise ratio (SNR) environments, the UAF converges to the identity function, which has near optimal root mean square error of $0.4888 \pm 0.0032$ $\mu M$. In the BipedalWalker-v2 RL dataset, the  UAF achieves the 250 reward in $961 \pm 193$ epochs, which proves that the UAF converges in the lowest number of epochs. Furthermore, the UAF converges to a new activation function in the BipedalWalker-v2 RL dataset.


\end{abstract}

\section{Introduction}\label{intro}

The goal of most machine learning tasks is to find the optimal model for a specific application. However, finding the optimal machine learning model by hand is a daunting task due to the virtually infinite number of possibilities on model and the corresponding parameter selection. The field of automated machine learning \cite{he2019automl,floreano2008neuroevolution,yao2018taking} was created to solve the problem, of which consists of automatically finding machine learning models using
genetic algorithms,  neural networks and its combination with probabilistic  and clustering algorithms.

Genetic algorithms are good at optimizing discrete variables. For example, they can be used to optimize the number of neurons in each layer or the number of  connections between each layer. NeuroEvolution of Augmenting Topologies (NEAT) \cite{stanley2002evolving} uses genetic algorithms to optimize the structure of neural networks. The weights of the neurons, the types of activation functions, and the number of neurons can be optimized by breeding and mutating different species of neural networks. HyperNEAT \cite{stanley2009hypercube} is a special neuroevolution algorithm. Instead of finding the architecture directly, it finds a single function that encodes the entire network. The single function is breed and  mutated in order to find the best function that encodes the optimal neural architecture. Deep HyperNEAT \cite{sosadeep} is an extension of HyperNEAT that allows for the design of larger neural networks.

Aside from genetic algorithms, neural network structures can also be optimized by other neural networks. Liu et al. \cite{liu2018progressive}
propose a new method for creating CNNs from scratch. CNNs are built from cells. Each cell does a specific operation such as activation function, convolution, concatenation, and pooling.  A predictor is trained to place cells and connect cells together. The CNN begins as a collection of a few cells and it grows as more cells are added.  The Efficient Neural Architecture Search via Parameter Sharing (ENAS) \cite{pham2018efficient} is similar because the RNN controller is trained using policy gradient to produce neural architectures using cell blocks. On the other hand, the Auto-DeepLab paper \cite{liu2019auto} proposes a method to search architectures at a cell level and at the network level.

Probabilistic methods use probability density functions (PDFs) to map between the input distributions and the output distributions. Given any input, the probabilistic methods predict the output values and the pertaining probability distribution of the output values. Moreover, probabilistic methods could be combined with neural network approaches to create new neural network architectures. Zoph et al. \cite{zoph2016neural} designed a recurrent neural network (RNN) controller for neural architecture search,  which was trained using reinforcement learning. The RNN controller searches through the vast array of possible neural networks and labels each network with a probability of being the optimal network. Moreover, it predicts the discrete parameters of the optimal network such as the size of the convolutional neural network (CNN) filters, number of CNN channels, and the type of activation function.

Clustering algorithms assign an identifier to each data point. Similar data points are clustered together based on the identifiers and the distance function. Clustering algorithms can be used to classify the type of problem based on the dataset. For example, the problem may be classified as a video quantification problem or a text classification problem or a reinforcement learning problem. Subsequently, the best neural network is selected from a pre-built model zoo and it is retrained to get the best results.

One of the core tasks for automated machine learning is to find an optimal activation function for a specific model. However, many activation functions have been proposed over the history of machine learning and this makes the selection difficult. The first neural network \cite{rosenblatt1958perceptron,rumelhart1986learning}  used the sigmoid activation function, where the outputs of the activation functions are limited to range $[0,1]$. The sigmoid function is good for limiting the outputs of neural networks. The sigmoid function belongs to the sigmoid activation function family, of which the general sigmoid equation \cite{richards1959flexible} was developed by F. J. Richards in 1959. For the most part, the sigmoid family is used for classifying objects, where $\hat{y}=1$ is the object existing and $\hat{y}=0$ is the object not existing. Other activation functions in the sigmoid family include the step function, the tanh function \cite{kalman1992tanh} and clipped function. Unlike the sigmoid function, the step function has a discontinuity at $x=0$ and only outputs $0$ or $1$. On the other hand, the tanh function is similar to the sigmoid function as the tanh function is constrained to range of $[-1,1]$.

The ReLU activation function \cite{hinton1997generative} is another popular activation function. The ReLU activation function outputs $y=0$ if $x < 0$, otherwise it outputs $y=x$. Moreover, the ReLU function is part of the ReLU activation function family, where the behaviour of all functions in the family are linear $y=x$ when $x>0$. The ReLU activation function family is mainly used for classification and reinforcement learning (RL) problems. The identity, LeakyReLU \cite{maas2013rectifier}, Elu \cite{clevert2015fast}, and softplus \cite{zheng2015improving} functions are included in this family. The identity function $y=x$ is typically used for the output layer of regression. The LeakyReLU is a version of ReLU that has a slight slope $y=\alpha x$ when $x<0$. The slight slope is used to prevent the gradient from reaching zero. One of the problems the ReLU and LeakyReLU functions encounter is the discontinuity at $x=0$ that produces undetermined gradients \cite{lu2019dying}. To remove undetermined gradients, the Elu, and softplus function are developed to have smoothness around $x=0$ \cite{zheng2015improving}.

The Gaussian activation function \cite{hartman1990layered} has a bell shaped curve and is useful for modeling Gaussian distributed random variables. For example, a neural network predicting the speed of a car might use the Gaussian function for regression because the speed of a car is Gaussian distributed \cite{noureldin2004ins}. Sometimes, the Gaussian function is used for classifying the existence of objects \cite{park1991universal}. The Gaussian function is a  special case of the radial basis functions (RBFs) \cite{park1991universal}, whose functions are always shaped like a bell shape curve.  Other members of the RBFs include the polyharmonic spline and the bump function. Newer activation functions such as Mish \cite{misra2019mish} and Swish \cite{ramachandran2017searching} have built-in regularization to prevent over-fitting of models. They look similar to the softplus and Elu activation functions, however they converge to $y=0$ when $x \to -\infty$ in order to eliminate large negative values \cite{misra2019mish,ramachandran2017searching}.

Among the many basic activation functions, selecting the best function that suits a specific task is hard. To overcome this problem, trainable parameters are added to the basic activation functions above. Subsequently, optimization algorithms are applied to find the best activation function parameters. PReLU \cite{xu2015empirical} is an example of adaptive activation function, where the slope $\alpha$ of a LeakyReLU function is a trainable parameter. Bodyanskiy et al. \cite{bodyanskiy2015evolving} develop an adaptable RBF that can be trained in real time. Qian et al. \cite{qian2018adaptive} propose adaptive ReLU functions for convolutional neural networks (CNNs). Campolucci et al. \cite{campolucci1996neural} propose an adaptive spline to approximate the curves of a sigmoid function. The uniformly sampled spline uses fixed knot vectors, fixed basis matrices, and the control points as the trainable parameters. The main problem with splines is over-fitting \cite{scardapane2017learning}. As the splines can fit all possible functions using the many trainable parameters, the spline may over-fit to the training set and may perform significantly worse in the testing test. Furthermore, the splines requires many additional constraints to allow continuity and differentiability.

To solve the problems of finding an optimal activation function, we propose a simple universal activation function (UAF) that can evolve to any of the above mentioned activation functions. Without any additional constraints, the UAF is continuous and differentiable for all parameter values. This enables the gradient descent algorithms to tune the UAF's parameters in order to achieve the best activation function for a specific problem. For example, the UAF can be initialized as the identity function. After training, the UAF might converge to the Mish function, which could be the optimal activation function for the particular problem. In other problems, the UAF might converge to LeakyReLU or Gaussian functions.

The paper is organized as follows. Section \ref{activationgen} describes UAF and its parameters. Section \ref{experiment} shows the UAF's performances on the CIFAR-10 \cite{krizhevsky2009learning} classification,  infrared spectra database for 9  gas quantification \cite{gan2019multi}, and BipedalWalker-v2 \cite{1606.01540} RL datasets. A conclusion is presented in Section \ref{con}. In the end, the appendix gives implementation details about the UAF.

\section{Universal Activation Function for Machine Learning }\label{activationgen}

The ReLU and softplus family of activation functions consist of monotonically increasing functions, which can be approximated by a softplus function given by
\begin{equation}
softplus(x)= \ln(1+e^{x})
\end{equation}
The softplus function can be generalized by adding parameters $A$ and $B$
\begin{equation}
    f_{UAF}(x) =\ln(1+e^{A(x+B)})
\end{equation} 
where parameter $B$ controls the horizontal shift.  For the LeakyReLU family of activation functions, they can be approximated by adding another monotonically decreasing function
\begin{equation}
f_{UAF}(x) =\ln(1+e^{A(x+B)}) - \ln(1+e^{D(x-B)} )
\end{equation}
and a new parameter $D$. This enables the slope $\alpha$ of the LeakyReLU family to be changed. Furthermore, the sigmoid and tanh family of activation functions can be approximated by
\begin{equation}
    f_{UAF}(x) =\ln(1+e^{A(x+B)}) - \ln(1+e^{D(x-B)} ) + E
\end{equation} 
adding a new parameter $E$ that shifts the entire function downwards to be centered at $y=0$ for the tanh approximation. To approximate the Gaussian family of activation functions
\begin{equation}
f_{UAF}(x) =\ln(1+e^{A(x+B)+Cx^{2}}) - \ln(1+e^{D(x-B)} ) + E 
\label{eq:UAFeq} 
\end{equation}
the parameter $C$ is added and thus the universal activation function $f_{UAF}(x)$ is now completed. In the supplementary materials, there is a video that describes the effect of the parameters on the UAF. It is evident that the UAF demonstrated in Eqn. \eqref{eq:UAFeq} is well behaved such that both the function and its first order derivative exist, are single valued and continuous for $x \in (-\infty,\infty)$ provided that all parameters are real.  In the following subsections, we will discuss in detail on how the UAF can be used to approximate various activation functions.

\subsection{Identity Function}
By setting $A=1$, $B=0$, $C=0$, $D=-1$, and $E=0$, the $f_{UAF}(x)$ becomes
\begin{equation}
f_{UAF}(x) = \ln \left (  \frac{1+e^{x}  }{  1+e^{-x}   }    \right ) =  x
\label{eq:identityuaf} 
\end{equation}
which is identical to the identity function.

\subsection{Step Function}
If $A=D\to \infty$, $B= \frac{1}{2A}$, $C=E=0$, then the $f_{UAF}(x)$ becomes
\[
f_{UAF}(x) = \lim_{A \to \infty}  \ln  \left ( \frac{  1+e^{Ax+0.5}  }{  1+e^{Ax-0.5}  }     \right )
\label{eq:stepuaf} 
\]
\begin{equation}
 =  \begin{cases}
    1 &  \text{ if } x > 0  \\
    0.5  &  \text{ if } x = 0  \\
    0  & \text{ otherwise }
 \end{cases}
\label{eq:step} 
\end{equation}
which is identical to the step function $f_{step}(x)$.

\begin{figure*}[!t]
\centering
\includegraphics[width=1\textwidth]{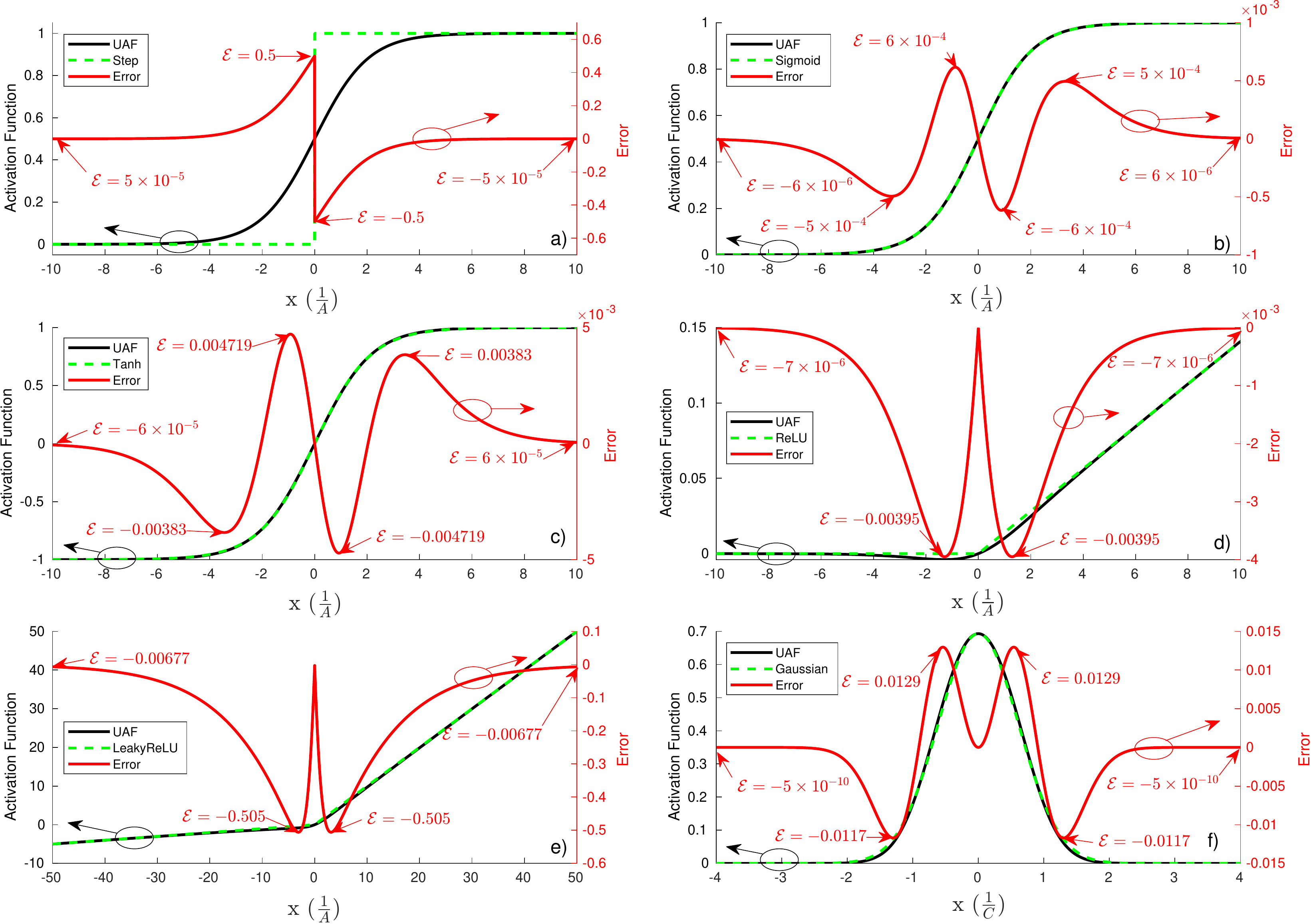}
\caption{ a) shows the step function approximation. b) shows the sigmoid function approximation. c) shows the tanh function approximation.  d) shows the ReLU function approximation. e) shows the LeakyReLU function approximation. f) shows the Gaussian function approximation. The black solid lines are the UAF, while the green dashed lines are targeted activation functions, whose values can be obtained from the $y$ axis on the left. The red solid lines represent the error ${\cal E}$ between the UAF and targeted activation function and the values can be read from the $y$ axis on the right side. }
\label{fig:UAFplot}
\end{figure*}

In practical implementations, $A$ is set to a finite large number that does not cause overflows. The error ${\cal E}$ between the $f_{UAF}(x)$ and the targeted function $f_{act}(x)$ (in this section, $f_{act}(x)=f_{step}(x))$ \begin{equation}
{\cal E}  = f_{UAF}(x) - f_{act}(x)
\label{eq:err} 
\end{equation}
can be analyzed as follows. 
At $x=0$, the
$f_{UAF}(0)  =   \ln  \left ( \frac{  1+e^{0.5}  }{  1+e^{-0.5}  }        \right ) 
=  0.5$ and this matches step function's value.  For values $Ax \gg 1$, $f_{UAF}(x) \approx 1.0$ as $e^{Ax\pm{0.5}}\gg1$ while for $Ax\ll -1$, $f_{UAF}(x) \approx 0$ since $e^{Ax{\pm}0.5}\ll1$. The derivative of the $f_{UAF}(x)$ is 
\begin{equation}
f_{UAF}'(x) 
= \frac{A e^{Ax} (e^{0.5}  - e^{-0.5} ) }{   (1+e^{Ax-0.5} ) (1+e^{Ax+0.5} )   } > 0 , \forall x \in \mathbb{R} 
\label{eq:stepderi} 
\end{equation}
Therefore, for $x > 0$, the error  ${\cal E}(x)<0$ suggesting the absolute error $|{\cal E}|$  monotonically decreases as $x$ increases in this regime. For $x<0$, ${\cal E}(x)>0$ and $|{\cal E}|$ montontonically increases as $x$ increases. The maximum absolute errors occur when x approaches zero from both sides. Due to the discontinuity of the step function, the error reaches $\pm{0.5}$ as $x\rightarrow 0^{\pm}$.

To quantitatively analyze the error between the UAF and the step function, we plot them for $x \in (-\frac{10}{A},\frac{10}{A})$. Here, $A=D=70.9992$, $B= 0.007042$, and $E=C=0$. As shown in Fig. \ref{fig:UAFplot}a), the solid black line represents the $f_{UAF}(x)$,   the dashed green line represents the step function, and the solid red line represents the error. Near $x=0$, the error is very high $|{\cal E}|= 0.5$. As $x$ moves away from $x=0$, the error decreases to $|{\cal E}|=5 \times 10^{-5}$ at $x=\pm \frac{10}{A}$. As a result, the absolute errors for points out of range are smaller than $5 \times 10^{-5}$. Clearly our UAF well approximates the step function for $x \in (-\infty,\infty)$.

\subsection{ Sigmoid Function}

As the step function is similar to the sigmoid function, the step function transforms into the sigmoid function, when $A = D \approx 1.01605291 $. The optimal value of $A$ is found using a gradient descent optimizer. The other parameters are setup exactly the same way as in the step function case with $B=\frac{1}{2A}=0.492100$, $C = E = 0$, written as
\begin{equation}
f_{UAF}(x) =  \ln \left (  \frac{  1+e^{Ax + 0.5 }  }{  1+e^{Ax - 0.5}  }     \right ) \end{equation}
and it gives a good approximation of the sigmoid  function $\sigma(x)$ given by
\begin{equation}
\sigma(x) = \frac{1}{ 1+e^{-x} }.
\end{equation}

The UAF approximation is bounded by the maximum absolute errors. In order to find the maximum absolute errors, the derivative of the error is set to zero $\frac{\mathrm{d} {\cal E} }{\mathrm{d} x}=0$ and the characteristic equation
\begin{equation}
(e - 1) A (e^x + 1)^2 e^{A x - 0.5} + e^x (e^{A x - 0.5} + 1) (-e^{A x + 0.5} - 1) = 0
\end{equation}
is solved to get the critical points located at
$
x \approx \pm 0.866499
$. The points have a maximum absolute error of 
\[
|{\cal E}|  =  \ln \left (  \frac{  1+e^{A(0.866499)+0.5 }  }{  1+e^{(0.866499)-0.5}  }     \right )  - \frac{1}{ 1+e^{-0.866499} }  
\]
\begin{equation}
\approx 0.000616
\end{equation}
for the approximation.

Similar to the case of the step function discussed above, at $x=0$, the
$f_{UAF}(0)  =   \ln  \left ( \frac{  1+e^{0.5}  }{  1+e^{-0.5}  }        \right ) 
=  0.5$ and this matches simoid function's value. As $x$ increases past zero, the absolute error increases up until the maximum absolute error $|{\cal E}| \approx 0.000616$ at $
x \approx  0.866499
$. As $x$ increases past that point, the absolute error monotonically decreases because the derivative is always positive $f_{UAF}'(x) > 0$ as mentioned in  \eqref{eq:stepderi} and the upper bound of the function $ f_{UAF}(x) = \sigma(x) \approx 1$, $ \forall Ax \gg 1 $. The inverse is also true. As $x$ decreases past the maximum error $
x \approx  -0.866499
$, the error monotonically decreases because the derivative is always positive $f_{UAF}'(x) > 0$ and the function has a lower bound $ f_{UAF}(x) = \sigma(x) \approx 0$, $ \forall Ax \ll -1$. As shown in Fig. \ref{fig:UAFplot}b), the solid black line represents the $f_{UAF}(x)$,   the dashed green line represents the sigmoid function, and the solid red line represents the error. Near $Ax=1$, the absolute error is high $|{\cal E}|= 6 \times 10^{-4}$. As $x$ moves away from the maximum absolute errors, the absolute error decreases to $|{\cal E}|=6 \times 10^{-6}$ at $x=\pm \frac{10}{A}$. This means the absolute error will always be less than $|{\cal E}|=6 \times 10^{-6}$ for values $x >  \frac{10}{A}$ or equivalently $x <  -\frac{10}{A}$.

\subsection{Tanh Function}

The tanh function is similar to the sigmoid function.  By setting $B=\frac{1}{A}$, $C=0$, $D=A$, $E=-1$ and optimizing the parameter $A$, the sigmoid function transforms into the tanh function. Parameter $B$ is twice as large because tanh has twice the range of sigmoid. For the best fit, set $A \approx 2.12616013$ and the UAF becomes

\begin{equation}
f_{UAF}(x) =  \ln \left (  \frac{  1+e^{Ax+1}  }{  1+e^{Ax-1}  }     \right ) - 1,
\end{equation}
an approximation of the $\tanh(x)$ function given by
\begin{equation}
\tanh(x) = \frac{e^{x} - e^{-x}}{ e^{x} + e^{-x} }.
\end{equation}

The UAF approximation is bounded by the maximum absolute errors. In order to find the maximum absolute errors, the derivative of the error is set to zero $\frac{\mathrm{d} {\cal E} }{\mathrm{d} x}=0$ and it leads to a characteristic equation given by
\begin{multline}
-Ae^{A x - 1} + A e^{A x + 1} - 2 A e^{A x + 2 x - 1} + 2 A e^{A x + 2 x + 1} \\
- A e^{A x + 4 x - 1} + A e^{A x + 4 x + 1} - 4 e^{A x + 2 x - 1} - 4 e^{A x + 2 x + 1} \\
- 4 e^{2 A x + 2 x}  - 4 e^{2 x} = 0.
\end{multline}
Numerically solving the above equation, we found four roots at $x=\pm 0.435499$, $x=\pm3.4355$. Among them, the maximum absolute error $|{\cal E}| 
\approx 0.004719$ occurs at $
x \approx \pm 0.435499
$. Note that for $x>3.4355$, the error derivative has no more zero crossing points, suggesting the error will monotonically decrease to zero from ${\cal E} = 0.00383$ at $x= 3.4355$. Similarly, for $x<-3.4355$, the error will monotonically approach zero from ${\cal E} =-0.00383$ at $x= -3.4355$. The change of error as a function of $x$ is clearly demonstrated as red solid line in Fig. \ref{fig:UAFplot}c). As shown, $|{\cal E}|=6 \times 10^{-5}$ at $x=\pm \frac{10}{A}$. This means the absolute error will always be less than $|{\cal E}|=6 \times 10^{-5}$ for values $x >  \frac{10}{A}$ or equivalently $x <  -\frac{10}{A}$.

\subsection{ReLU Function}

If $B=C=E=0$, $D=A-1$,  $A \to \infty$, the $f_{UAF}(x)$ becomes
\[
f_{UAF}(x) =  \lim_{A \to \infty}   \ln  \left (  \frac{  1+e^{Ax}  }{  1+e^{Ax-x}  }    \right ) 
\]
\begin{equation}
=  \begin{cases}
x, & x \ge 0\\ 
0, & x <  0
\end{cases}
\label{eq:relu} 
\end{equation}
which is identical to the $ReLU(x)$ function.

In practical implementations, $A$ is set to a large finite number that avoids overflows. Under such circumstances, the maximum absolute errors  occur where the derivative of error becomes zero. For positive interval $x>0$, this leads to a characteristic equation
\begin{equation}
(A-1) e^{Ax} - A e^{(A-1)x} - 1 = 0.
\end{equation}
The maximum absolute error occurs at position $x \approx \ln \left ( \frac{  A}{A-1} \right ) \approx 0.0181$ and the value of the maximum absolute error is
\begin{equation}
|{\cal E}| \approx   \left |
\ln  \left (  \frac{  1+(\frac{A}{A-1})^{A}  }{  1+(\frac{A}{A-1})^{A-1}  }    \right )
-x \right | \approx 0.00395.
\end{equation}
For the negative interval $x < 0$, this leads to a characteristic equation
\begin{equation}
A e^x + e^{A x} + 1 = A.
\end{equation}
The maximum absolute error occurs at position
$x \approx \ln \left ( \frac{  A-1}{A} \right ) \approx -0.0181 $
and the value of the maximum absolute error is
\begin{equation}
|{\cal E}| \approx   \left |
\ln  \left (  \frac{  1+(\frac{A-1}{A})^{A}  }{  1+(\frac{A-1}{A})^{A-1}  }    \right )
\right | \approx 0.00395.
\end{equation}

At $x=0$, the
$f_{UAF}(0)  =   \ln  \left ( \frac{  1+e^{0}  }{  1+e^{0}}          \right ) 
=  0$ and this matches ReLU function's value. As $x$ increases past zero, the absolute error increases up until the maximum absolute error $|{\cal E}| \approx 0.00395$ at $ x \approx 0.0181$. As $x$ increases past that point, the absolute error monotonically decreases and converges to zero. For the negative interval $x < 0$, as $x$ decreases past the maximum absolute error point $
x \approx -0.0181
$, the absolute error monotonically decreases and  converges to zero. Fig. \ref{fig:UAFplot}d) shows a red line that represents the error of the UAF. Around the neighbourhood $x=0$, the error is high. As $x$ moves away from the neighbourhood around $x=0$, the error decreases to $|{\cal E}|=7 \times 10^{-6}$ at $x=\pm \frac{10}{A}$. This means the absolute error will always be less than $|{\cal E}|=7 \times 10^{-6}$ for values $x >  \frac{10}{A}$ or equivalently $x <  -\frac{10}{A}$.

\subsection{ LeakyReLU Function}

LeakyReLU functions have a parameter $\alpha \in (0,0.1]$ that controls the negative slope. By setting the parameters $A=1$, $B=C=E=0$, $D=-\alpha$,
\begin{equation}
f_{UAF}(x) =  \ln \left (  \frac{  1+e^{x }  }{  1+e^{-\alpha x}  }     \right ) 
\end{equation}
gives an approximation of the LeakyReLU function 
\begin{equation}
LeakyReLU(x) =  \begin{cases}
x, & x \ge 0\\ 
\alpha x, & x <  0
\end{cases}
\end{equation}

The maximum absolute errors  occur where the derivative of error $ \frac{\mathrm{d} {\cal E} }{\mathrm{d} x} = 0 $. For positive interval $x>0$, this leads to a characteristic equation
\begin{equation}
(\alpha-1) e^{-\alpha x} + \alpha e^{x - \alpha x}  - 1 = 0.
\end{equation}
For negative interval $x<0$, this leads to a characteristic equation.
\begin{equation}
\alpha (-e^x - 1) + e^x (e^{-\alpha x} + 1) = 0.
\end{equation}
If $\alpha=0.1$, the maximum absolute error is $|{\cal E}|=0.505$ at position $x= \pm 3.106$. At $x=0$, $f_{UAF}(0)  =   \ln  \left ( \frac{  1+e^{0}  }{  1+e^{0}}          \right ) 
=  0 $ and this matches the LeakyReLU's value. As $x$ decreases below 0, the absolute error increases up until the maximum absolute error. After decreasing past the point $x= -3.106$, the absolute error given by 
\begin{equation}
{\cal E}(x) = \ln  \left (  \frac{  1+e^{x}  }{  1+e^{-\alpha x}  }    \right ) - \alpha x  = \ln  \left (  \frac{  1+e^{x}  }{  1+e^{\alpha x}  }    \right )
\end{equation}
monotonically decreases and converges to zero $\lim_{x \to -\infty} {\cal E}(x) =0$. The inverse is also true. As $x$ increases above 0, the absolute error increases up until the maximum absolute error. After increasing past the point $x= 3.106$, the absolute error given by
\begin{equation}
{\cal E}(x) = \ln  \left (  \frac{  1+e^{x}  }{  1+e^{-\alpha x}  }    \right ) -  x  = \ln  \left (  \frac{  1+e^{-x}  }{  1+e^{-\alpha x}  }    \right ) 
\end{equation}
monotonically decreases and converges to zero $\lim_{x \to \infty} {\cal E}(x) =0$. Fig.~\ref{fig:UAFplot}e) represents the error of the UAF as a red line. Around the neighbourhood $x=0$, the error is very high $|{\cal E}|=0.505$. As $x$ moves away from $x=0$, the error decreases to $|{\cal E}|=0.00677$ at $x=\pm \frac{50}{A}$. This means the absolute error will always be less than $|{\cal E}|=0.00677$ for values $x >  \frac{50}{A}$ or equivalently $x <  -\frac{50}{A}$.

\subsection{Softplus Function}
If $A=1$, $B=0$, $C=0$, $D=0$ and $E=\ln(2)$, the $f_{UAF}(x)$ becomes
\[
f_{UAF}(x) =  \ln \left (  \frac{  1+e^{1(x+0)+0x^{2}}  }{  1+e^{0(x-0)}  }     \right ) + \ln(2)
\]
\begin{equation}
=  \ln ( 1+e^{x}  )
\end{equation}
which exactly equals the softplus function because the UAF is built using the softplus function.

\subsection{Gaussian Function}

A gradient descent optimizer is used to find the parameters $C=-0.61341425$ and $E=\ln(2)$ such that the UAF with the form
\begin{equation}
f_{UAF}(x) =  \ln  (   1+e^{-0.61341425x^{2}}     ) 
\end{equation}
approximates a scaled version of the Gaussian function $H(x)$
\begin{equation}
H(x) = \ln(2) e^{  -\frac{x^{2}}{2}  } 
\end{equation}
with minimum RMSE over $x \in (-\infty,\infty)$. The other parameters are not needed and they are set to zero $A=B=D=0$.  The maximum absolute error occurs when the derivative of the absolute error is zero, which leads to the characteristic equation 
\begin{equation}
\frac{2Cx e^{Cx^{2}} }{ 1+ e^{Cx^{2}} }   + x \ln(2) e^{  -\frac{x^{2}}{2}  }  = 0
\end{equation}
and the critical points occur at $x\approx \pm 0.8821$ and has a value of $|{\cal E}| = 0.0129$

At $x=0$, the
$f_{UAF}(0)  =   \ln  \left ( 1 + e^0          \right ) 
=  \ln(2)$ and this matches Gaussian function's value $N(0) = \ln(2)$. As $x$ increases past zero, the absolute error increases up until the maximum absolute error $|{\cal E}| \approx 0.0129$ at $ x \approx 0.8821$. As $x$ increases past that point $-Cx \gg 4$, the absolute error converges to zero because the UAF goes to zero $f_{UAF}(x) \approx 0$ and the Gaussian function goes to zero $N(x) \approx 0$. The inverse is also true.  Fig. \ref{fig:UAFplot}f) shows a red line that represents the error of the UAF. Around the neighbourhood $x=0$, the error is high $|{\cal E}|=0.0129$. As $x$ moves away from $x=0$, the absolute error exponentially decreases to $|{\cal E}|=5 \times 10^{-10}$ at $x=\pm \frac{4}{C}$. This means the absolute error will always be less than $|{\cal E}|=5 \times 10^{-10}$ for values $x >  -\frac{4}{C}$ or equivalently $x <  \frac{4}{C}$.

\subsection{UAF RMSE Table}

\begin{table}[h!]
\caption{ \label{tab:rmseuaf} RMSE of Various Activation Functions.} 
\begin{center}
    \begin{tabular}{l l l l }
    \hline
\hline
Activation Function & RMSE  in $[-10,10]$  & Max Error & Error Locations   \\ \hline
Identity            & 0.00000    &  0.00000 & None  \\ \hline
Step                & 0.01664 &  0.50000 & $\epsilon$\\ \hline
ReLU                & 0.00021  & 0.00395 & $\pm 0.0181$\\ \hline
LeakyReLU           & 0.41316  & 0.50501 & $\pm3.1064$\\ \hline
Sigmoid             & 0.00029 & 0.00062 & $\pm 0.8665$\\ \hline
Tanh                & 0.00160 & 0.00472 & $\pm0.4355$\\ \hline
Softplus            & 0.00000  &  0.00000  &  None  \\ \hline
Gaussian            & 0.00468   & 0.0129 &  $\pm0.8821$ \\ \hline
    \end{tabular}
\end{center}
\end{table}

To further compare the goodness of the UAF approximation, Table \ref{tab:rmseuaf} shows the RMSEs for each activation function in the interval of $x \in [-10,10]$. The UAF models the Identity function and the softplus function perfectly. For the continuous activation functions such as the sigmoid, tanh, and Gaussian, the UAF models them well with a small RMSE. For the discontinuous activation functions, the RMSE is slightly higher due to the UAF not being able to mimic the discontinuities.

\section{Experiments} \label{experiment}

In this article, three experiments are used to benchmark the different activation functions including the UAF. Starting from the easiest dataset, CIFAR-10 \cite{krizhevsky2009learning} is used to benchmark image classification, which many researchers have achieved $F_{1} > 0.9$ with ensembles of  CNNs \cite{kolesnikov2019big,huang2019gpipe,tan2019efficientnet}.

\begin{table}[h!]
\caption{ \label{tab:VGG8cifar10} CIFAR-10 : VGG 8 Layers.} 
\begin{center}
    \begin{tabular}{ p{1.2 cm} p{1cm} p{1cm} p{1cm} p{1cm} p{0.8cm}}
    \hline
\hline
Activation Function & Precision & Recall & $F_{1}$ & Training Time (s) & Batches \\ \hline

Identity & $0.8038 \pm 0.0089$
& $0.7980 \pm 0.0149$
& $0.7952 \pm 0.0172$
& $656 \pm 5.6$
& $10000$ \\ \hline

ReLU & $0.0100 \pm 0.0010$
& $0.1000100$
& $0.0182 \pm 0.0010$
& $650 \pm 4.5$
& $10000$ \\ \hline

LeakyReLU & $0.8931 \pm 0.0033$
& $0.8931 \pm 0.0032$
& $0.8930 \pm 0.0033$
& $863 \pm 4.8$
& $10000$ \\ \hline

Sigmoid & $0.8820 \pm 0.0043$
& $0.8806 \pm 0.0059$
& $0.8807 \pm 0.0056$
& $686 \pm 10$
& $10000$ \\ \hline

Tanh & $0.8390 \pm 0.0061$
& $0.8352 \pm 0.0107$
& $0.8350 \pm 0.0097$
& $703 \pm 5.6$
& $10000$ \\ \hline

Softplus & $0.9024 \pm 0.0028$
& $0.9021 \pm 0.0026$
& $0.9021 \pm 0.0026$
& $699 \pm 1.8$
& $10000$ \\ \hline

ELU & $0.8858 \pm 0.0042$
& $0.8858 \pm 0.0042$
& $0.8856 \pm 0.0043$
& $699 \pm 3.2$
& $10000$ \\ \hline

Mish & $0.8909 \pm 0.0079$
& $0.8903 \pm 0.0082$
& $0.8905 \pm 0.0081$
& $927 \pm 5.1$
& $10000$ \\ \hline

UAF & $0.9019 \pm 0.0039$
& $0.9016 \pm 0.0041$
& $0.9017 \pm 0.0040$
& $1692 \pm 4.2$
& $10000$ \\ \hline

    \end{tabular}
\end{center}
\qquad \qquad Note: Macro averaged results. Confidence interval of $2\sigma$.
\end{table}

\begin{figure}[!t]
\centering
\includegraphics[width=0.5\textwidth]{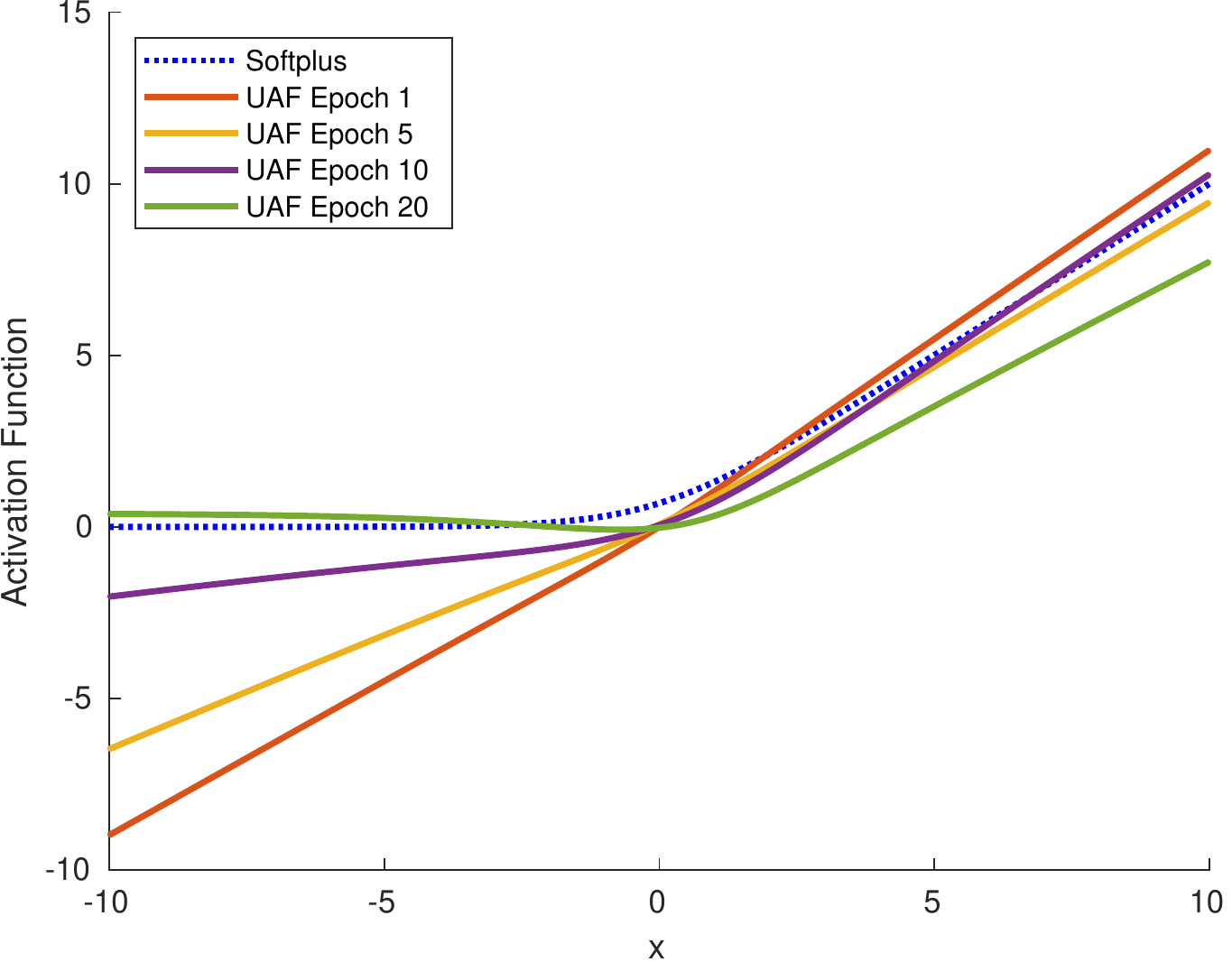}
\caption{ CIFAR 10 VGG 8 layer UAF evolution. }
\label{fig:uafcifar}
\end{figure}

 CIFAR-10 contains many 32x32x3 images and the goal is to classify airplanes, automobiles, birds, cats, deers, dogs, frogs, horses, ships, and trucks. The VGG 8 layer CNN \cite{simonyan2014very} is applied to the CIFAR-10 dataset, which contains many CNN layers and max pooling layers. The activation functions remain constant across all layers and all neurons. The precision, recall, and $F_1$ scores measure the classification performances.
 Table \ref{tab:VGG8cifar10} shows the 1x10 fold performances of the different activation functions on the image classification dataset CIFAR-10 \cite{krizhevsky2009learning}. The ReLU activation function has the worst  $F_1$ score because the gradient gets stuck at zero and the weights do not update. The identity, sigmoid, tanh, and ELU activation functions have poor $F_1$ scores because their gradients do not back-propagate well across many different CNN layers. On the other hand, Mish and LeakyReLU functions are designed to preserve the gradient across many different CNN layers. As a result, they perform better than the sigmoid and ELU. Softplus and UAF have the highest $F_{1}$ score due to smoothness of the functions and being able to reach the global minimum. This means softplus and UAF are superior at classifying objects when compared to the other activation functions. However, the UAF requires more training time due to the complexity of the function. Fig. \ref{fig:uafcifar} shows the evolution of the UAF on the CIFAR-10 dataset. The UAF is initialized with the identity function.  As UAF is trained, the UAF converges to a Mish function that is shifted to the right and has a different slope. A corresponding animation is enclosed as supplementary information.

\begin{table}[h!]
\caption{ \label{tab:9gasmlp} 9 Gas Quantification : 2 Layer MLP.} 
\begin{center}
    \begin{tabular}{cccc }
    \hline
\hline
Activation Function & RMSE ($\mu$M) & Training Time (s) \\ \hline

Identity & $0.4891 \pm 0.0035$ & $2840 \pm 180$ \\ \hline

ReLU & $1.2 \pm 1.7$ & $2726 \pm 54$
  \\ \hline

LeakyReLU & $0.4879 \pm 0.0037$ & $2870 \pm 160$
  \\ \hline
  
Sigmoid & $0.9514 \pm 0.0094$ & $2810 \pm 140$ \\ \hline

Tanh & $0.6942 \pm 0.0016$ & $2890 \pm 110$ \\ \hline

Softplus & $0.897 \pm 0.031$ & $2840 \pm 160$ \\ \hline

ELU & $0.4900 \pm 0.0031$ & $2870 \pm 160$ \\ \hline

Mish & $0.5269 \pm 0.0043$ & $2900 \pm 130$ \\ \hline

UAF & $0.4888 \pm 0.0032$ & $2932 \pm 110$ \\ \hline

    \end{tabular}
\end{center}
\qquad \qquad Note: Infrared spectra database for 9  gas quantification \cite{gan2019multi} 30 dB SNR uniformly distributed concentrations. Confidence interval of $2\sigma$.
\end{table}

\begin{figure}[!t]
\centering
\includegraphics[width=0.5\textwidth]{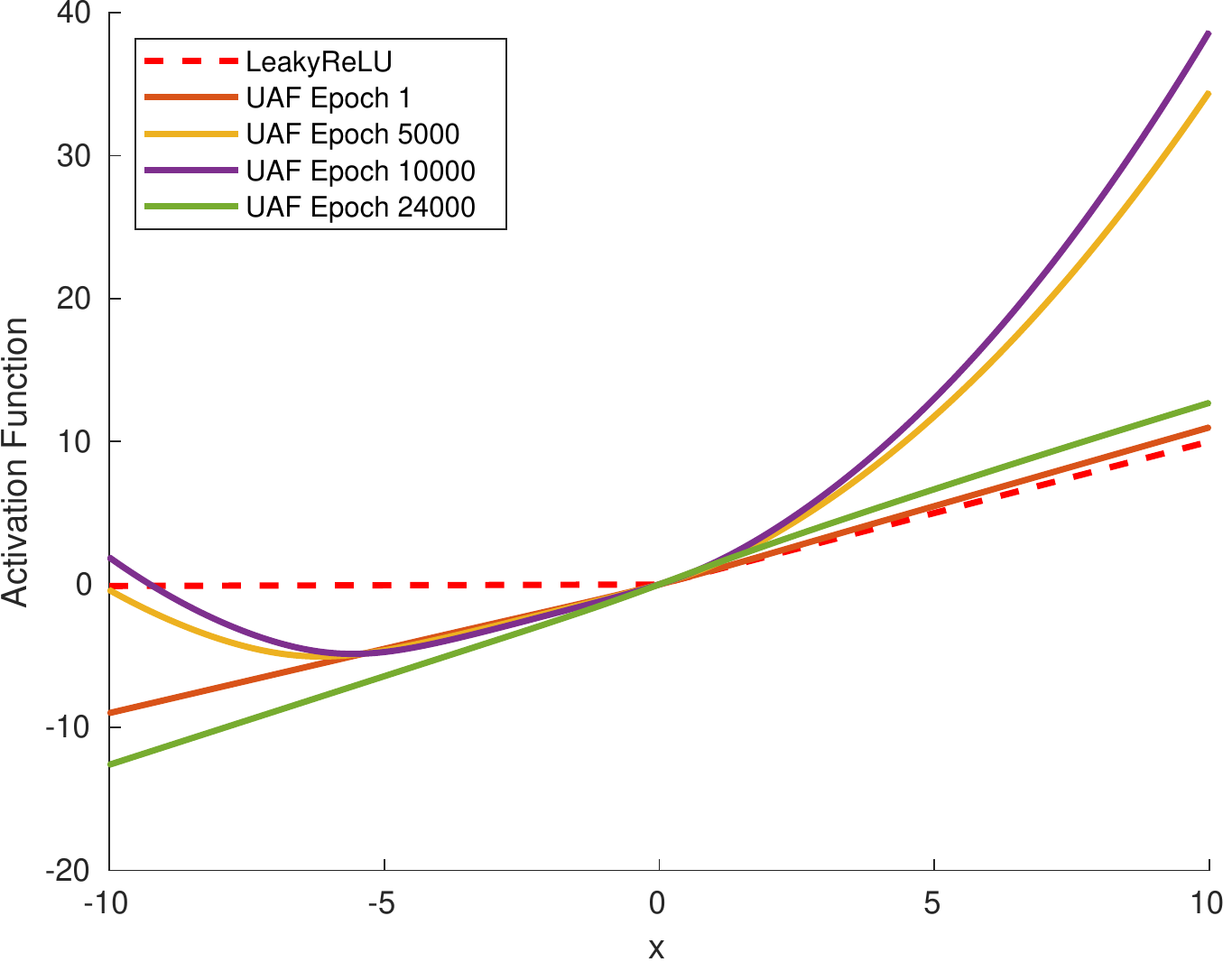}
\caption{ Infrared spectra database for 9  gas quantification \cite{gan2019multi} 30 dB SNR uniformly distributed concentrations UAF evolution. }
\label{fig:uaf9gas}
\end{figure}

The second hardest dataset is the infrared spectra database for 9  gas quantification. This current dataset is similar to the dataset used in \cite{gan2019multi} except that the gas concentrations are uniformly distributed and all gas concentrations are above zero. Quantification is slightly harder than classification because the network needs to predict a range of values instead of a binary result. 1D spectra of various gasses are fed into a 2 layer MLP, of which the activation functions remain constant for all layers. The MLP predicts the concentrations of 9 gases using 109 neurons. Table \ref{tab:9gasmlp} shows the 1x10 fold testing of the infrared spectra database \cite{gan2019multi} with 30 dB SNR. The ReLU function sometimes gets stuck because the gradient can be zero. This causes the ReLU function to have a high RMSE. The softplus, sigmoid, and tanh functions have high RMSEs because they are not suited for quantification.  MLPs using the Identity, LeakyReLU, and UAF obtained the lowest RMSE. As a result, MLPs with the identity, LeakyReLU, and UAF are able to predict the concentrations of the gasses more accurately than the MLP with other activation functions. Fig. \ref{fig:uaf9gas} shows the evolution of the UAF during the training procedure. The UAF begins as the identity function. Afterwards, the UAF changes to a parabolic function. Subsequently, the UAF converges to the identity function, which is close to the optimal activation function. An animation depicting the evolution of the UAF is available in the supplementary materials.

\begin{table}[h!]
\caption{ \label{tab:bipedalwalker} BipedalWalker-v2 : Deep Deterministic Policy Gradient.} 
\begin{center}
    \begin{tabular}{ p{1.2 cm} p{1cm} p{1cm} p{1cm} p{1cm} p{0.8cm}}
    \hline
\hline
Activation Function & 100 Reward (Epoch) & 250 Reward (Epoch) & 40 Distance (Epoch) & 88 Distance (Epoch) & Training Time (s) \\ \hline

Identity
& $\infty$
& $\infty$
& $1378 \pm 363$
& $1491 \pm 52$
& $2501 \pm 1290$
\\ \hline

ReLU 
& $890 \pm 158$
& $991 \pm 191$
& $797 \pm 165$
& $831 \pm 177$
& $5532 \pm 863$
\\ \hline

LeakyReLU
& $850 \pm 159$
& $990 \pm 158$
& $773 \pm 144$
& $817 \pm 154$
& $6135 \pm 1719$
\\ \hline

Sigmoid 
& $818 \pm 213$
& $1004 \pm 413$
& $717 \pm 229$
& $792 \pm 265$
& $5933 \pm 1979$
\\ \hline

Tanh
& $919 \pm 201$
& $1223 \pm 481$
& $831 \pm 147$
& $896 \pm 198$
& $4461 \pm 2266$
\\ \hline

Softplus 
& $836 \pm 527$
& $992 \pm 592$
& $676 \pm 330$
& $754 \pm 290$
& $4937 \pm 2691$
\\ \hline

ELU 
& $928 \pm 164$
& $1173 \pm 453$
& $824 \pm 218$
& $874 \pm 127$
& $5855 \pm 2045$
\\ \hline

Mish & $994 \pm 233$
& $1293 \pm 252$
& $888 \pm 122$
& $939 \pm 131$
& $6954 \pm 2024$ \\ \hline

UAF
& $859 \pm 209$
& $961 \pm 193$
& $766 \pm 128$
& $832 \pm 193$
& $6983 \pm 2375$
\\ \hline

    \end{tabular}
\end{center}
\qquad \qquad Note: Confidence interval of $2\sigma$.
\end{table}

\begin{figure}[!t]
\centering
\includegraphics[width=0.5\textwidth]{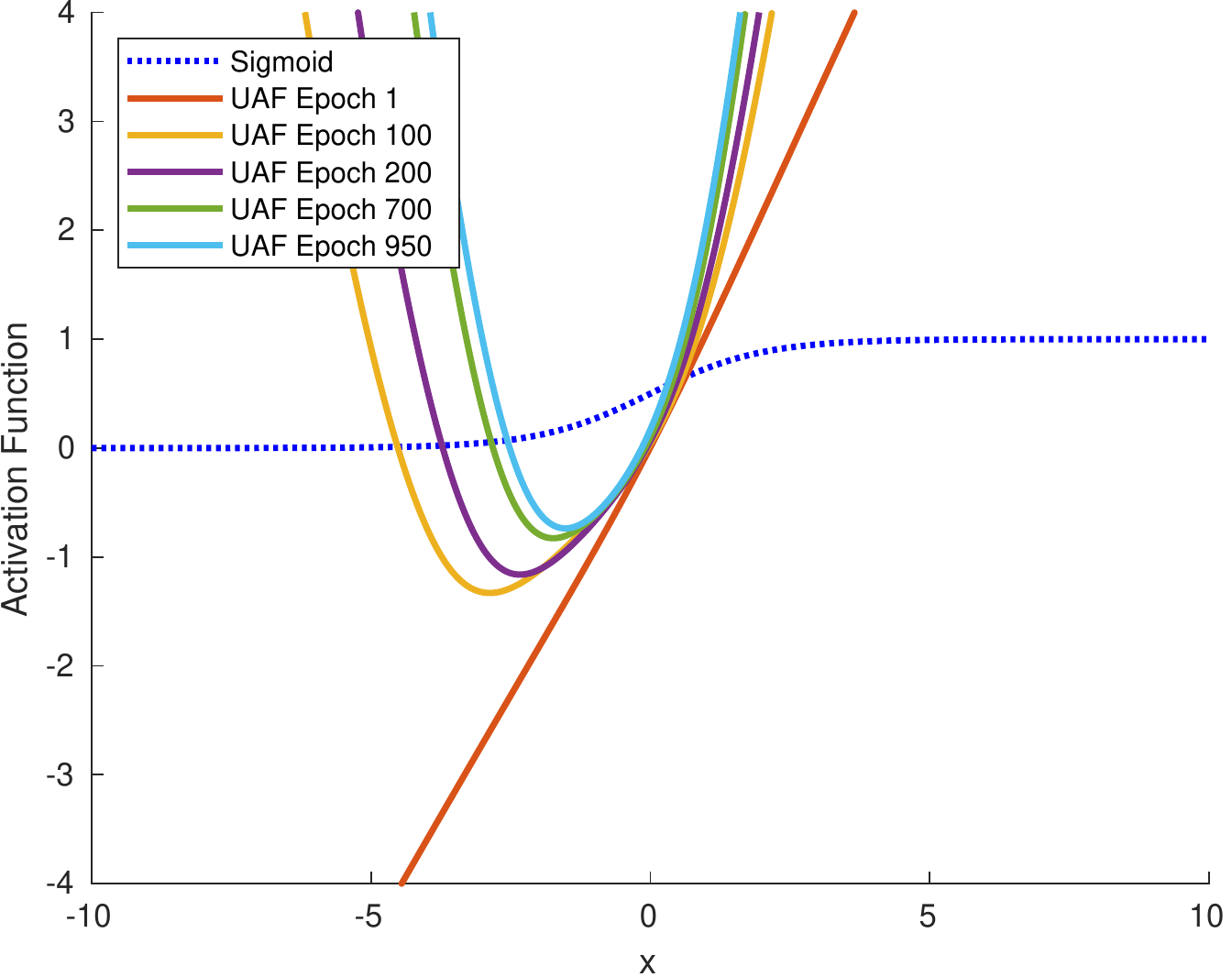}
\caption{ BipedalWalker-v2 Deep Deterministic Policy Gradient  UAF evolution. }
\label{fig:uafddpg}
\end{figure}

The hardest problem is the BipedalWalker-v2 \cite{1606.01540} reinforcement learning benchmark. The goal is to make the bipedal robot walk towards the finish line while the terrain of the simulation changes. The neural networks control the torques of the robot's legs in order to keep the robot from falling over. The reward function depends on the furthest distance traveled and the amount of torque applied. Further distances mean higher rewards and smaller torques mean higher rewards too. Moreover, the neural networks must converge in the least amount of epochs. High rewards and low number of epochs imply that the algorithms run efficiently.
Table \ref{tab:bipedalwalker} shows Deep Deterministic Policy Gradient \cite{ddpg} on BipedalWalker-v2 with different activation functions. As shown, the
sigmoid activation function achieves the 100 reward in $818 \pm 213$ epochs, which is the least amount of epochs. UAF is slightly slower in achieving the 100 reward with $859 \pm 209$ epochs. However, UAF  is the fastest at achieving the 250 reward with $961 \pm 193$ epochs. In the long run, the UAF achieves the best performance in terms of the rewards and the number of epochs.

Fig. \ref{fig:uafddpg} shows the evolution of the UAF in BipedalWalker-v2. The UAF is initialized as the identity function. Subsequently, the UAF evolves to an unusual parabolic activation function. The parabolic function is a new activation function that performs well for this specific problem. It limits the torque of the bipedal robot to $y \in [-1,\infty)$ and the parabolic function decreases the energy needed to move the robot. As the energy needed decreases, the reward increases. An animation depicting the evolution of the UAF is available in the supplementary materials.

\section{Conclusion and Future Work} \label{con}

A UAF has been developed to approximate every other activation function found in the literature. The UAF has parameters $A,B,C,D,E$ and is continuous for all parameter values. By adjusting the UAF's parameters, the UAF can evolve to step, identity, ReLU, LeakyReLU, sigmoid, tanh, Gaussian, or softplus. It has been shown that in the CIFAR-10 image classification, the UAF and the softplus function both achieve the highest $F_{1}$ score, which means they are close to optimal for CIFAR-10.  In the infrared spectra database for 9 gas quantification, the UAF along with LeakyReLU and identity have the lowest RMSE. As a result, they perform the best at gas quantification.  In the study of BipedalWalker-v2, the sigmoid function achieves the 100 reward in the least amount of epochs, while the UAF converging to a parabolic activation function achieves the 250 reward in the least amount of epochs. This means the UAF converged the fastest in terms of long term goals. In conclusion, incorporating the developed UAF in a neural network leads to near optimal performance, without the need to try many different activation functions in the design.

As for the future work, the UAF could be extended to 
other sets of numbers. For example, the UAF could be modified to take in complex numbers and this  allows for the application of the UAF to the complex number neural networks. In this paper, a single UAF is applied to the entire network. Instead of that, the UAF could be applied to individual layers or to individual neurons. This would enable the neural networks to model more non-linear processes. Moreover, the UAF could be used for transfer learning. The activation functions from one neural network could be transferred to any other neural network. This would enable neural networks to learn from each other and converge faster.

\begin{appendices}

\section{Implementation Details}

Batch normalization $BN(x)$ is widely used by researchers
\begin{equation}
BN(x) =  \frac{x - \mu_{x} }{ \sigma_{x} }
\label{eq:bn}
\end{equation}
where $\mu_{x}$, $\sigma_{x}$ are the mean and the standard deviation of input $x$ respectively. Batch normalization is normally applied to the data before the activation function. This effectively reduces the input $x$ domain of UAF the to $x \in [-100,100]$, which reduces floating point overflows. The softplus function \cite{tensorflow} can be rewritten to
\begin{equation}
\ln(1+e^{x}) = ReLU(x) + log1p(e^{-|x|})
\end{equation}
in order to increase the floating point precision. The UAF can be rewritten to 
\begin{equation}
\begin{split}
f_{UAF}(x) = ReLU(A(x+B)+Cx^{2}) \\ 
+ log1p(e^{-|A(x+B)+Cx^{2}|}) \\
+ ReLU(D(x-B))  \\ 
+ log1p(e^{-|D(x-B)|}) \\
+ E
\end{split}
\label{eq:UAF2}
\end{equation}
and this minimizes the floating point overflow errors.

\end{appendices}



\end{document}